\documentclass[10pt,twocolumn,letterpaper]{article}

\usepackage{cvpr}              

\usepackage[dvipsnames]{xcolor}

\usepackage{multirow}
\usepackage{amsmath}
\usepackage{graphicx}
\usepackage{algorithm}
\usepackage{algorithmic}
\usepackage{subcaption}
\usepackage{cuted}
\definecolor{color1}{rgb}{0.94,0.94,1}
\definecolor{color2}{rgb}{0.9,0.97,1}
\definecolor{color3}{rgb}{1,0.5529,0.5529}
\definecolor{color4}{rgb}{0.6902,0.7686,0.8706}
\definecolor{color5}{rgb}{1, 0, 0}

\definecolor{cvprblue}{rgb}{0.21,0.49,0.74}

\usepackage[pagebackref,breaklinks,colorlinks,citecolor=cvprblue]{hyperref}

\def\paperID{1685} 
\def\confName{CVPR}
\def\confYear{2024}

\title{Single-View Scene Point Cloud Human Grasp Generation}

\author{Yan-Kang Wang$^{1}$~~Chengyi Xing$^2$~~Yi-Lin Wei$^1$~~Xiao-Ming Wu$^1$~~Wei-Shi Zheng$^{1,3}$\thanks{Corresponding author}\\
$^1$School of Computer Science and Engineering, Sun Yat-sen University, China \\
$^2$Stanford University, Stanford, CA\\
$^3$Key Laboratory of Machine Intelligence and Advanced Computing, Ministry of Education, China \\
\tt\small \{wangyk65, weiylin5, wuxm65\}@mail2.sysu.edu.cn~~chengyix@stanford.edu wszheng@ieee.org
}

\begin{document}

{\onecolumn
\noindent \vspace{1cm}

\noindent \textbf{\huge\centering{Single-View Scene Point Cloud Human Grasp Generation}}

\vspace{2cm}

\noindent {\LARGE{Yan-Kang Wang, Chengyi Xing, \\
Yi-Lin Wei, Xiao-Ming Wu, Wei-Shi Zheng*}}
\\
\\
*Corresponding author: Wei-Shi Zheng.
\\
\\
Code: \href{https://github.com/iSEE-Laboratory/S2HGrasp}{\textcolor{blue}{https://github.com/iSEE-Laboratory/S2HGrasp}}
\\
\\
Project page: \href{https://isee-laboratory.github.io/S2HGrasp/}{\textcolor{blue}{https://isee-laboratory.github.io/S2HGrasp/}}

\vspace{1cm}

\noindent {\LARGE{Submission date: 27-Feb-2024 to  IEEE/CVF Conference on Computer Vision and Pattern Recognition}}

\vspace{1cm}

\noindent For reference of this work, please cite:

\vspace{1cm}
\noindent Yan-Kang Wang, Chengyi Xing, Yi-Lin Wei, Xiao-Ming Wu, Wei-Shi Zheng. ~Single-View Scene Point Cloud Human Grasp Generation.~~In \emph{Proceedings of the IEEE Conference on
Computer Vision and Pattern Recognition,} 2024.

\vspace{1cm}

\noindent Bib:\\
\noindent @inproceedings\{wang2024single,\\
\ \ \  title     = \{Single-View Scene Point Cloud Human Grasp Generation\}, \\
 \ \ \   author    = \{Wang, Yan-Kang and Xing, Chengyi and Wei, Yi-Lin and Wu, Xiao-Ming and Zheng, Wei-Shi\},\\
\ \ \  booktitle   = \{Proceedings of the IEEE/CVF Conference on Computer Vision and Pattern Recognition\},\\
\ \ \  year      = \{2024\}\\
\}
}
\twocolumn


\maketitle
\begin{strip}\centering
\vspace{-1.4cm}
\includegraphics[width=1\linewidth]{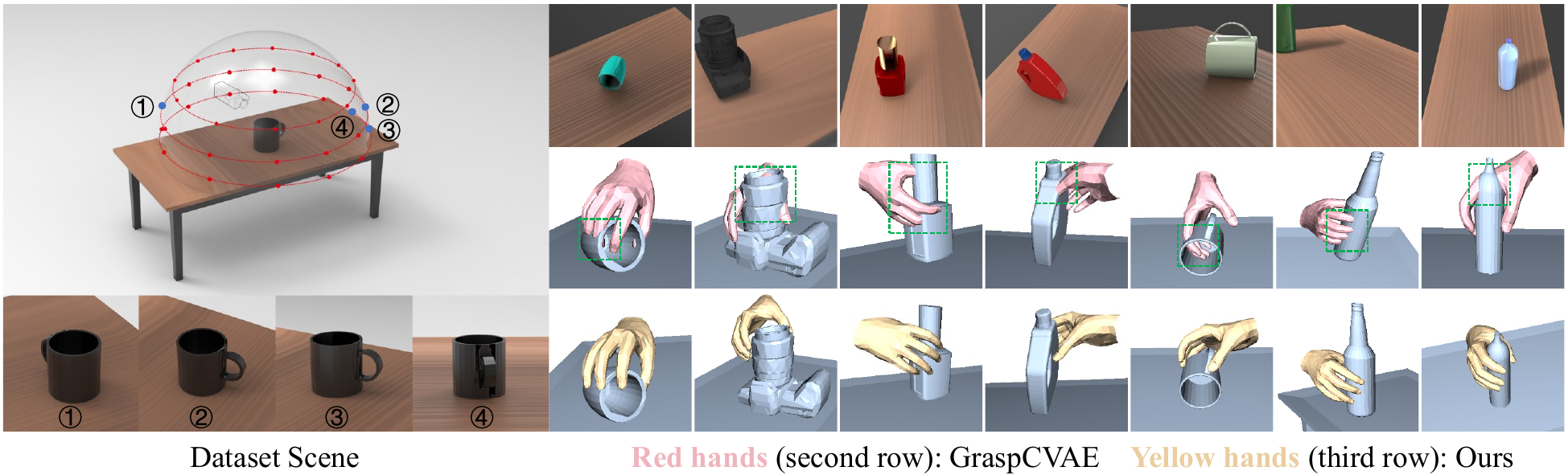}
\vspace{-0.7cm}
\captionof{figure}{The \textbf{scene} of our dataset and \textbf{comparison} between our model and GraspCVAE \cite{jiang2021hand}. The top-left image depicts the scene in our dataset, comprising a table and an object. Below it are images from four random viewpoints out of $36$. The three rows of images on the right side represent, respectively, the single-view images, the generation results of GraspCVAE and our method. The green box indicates the area where the hand-object penetration occurs due to the lack of global perception ability of GraspCVAE.
\label{dataset and results}}
\vspace{-.2cm}
\end{strip}
\begin{abstract}
In this work, we explore a novel task of generating human grasps based on single-view scene point clouds, which more accurately mirrors the typical real-world situation of observing objects from a single viewpoint. Due to the incompleteness of object point clouds and the presence of numerous scene points, the generated hand is prone to penetrating into the invisible parts of the object and the model is easily affected by scene points.
Thus, we introduce \textbf{S2HGrasp}, a framework composed of two key modules: the Global Perception module that globally perceives partial object point clouds, and the DiffuGrasp module designed to generate high-quality human grasps based on complex inputs that include scene points. Additionally, we introduce \textbf{S2HGD} dataset, which comprises approximately 99,000 single-object single-view scene point clouds of 1,668 unique objects, each annotated with one human grasp. Our extensive experiments demonstrate that S2HGrasp can not only generate natural human grasps regardless of scene points, but also effectively prevent penetration between the hand and invisible parts of the object. Moreover, our model showcases strong generalization capability when applied to unseen objects. Our code and dataset are available at \textcolor{magenta}{https://github.com/iSEE-Laboratory/S2HGrasp}.
\end{abstract}
\section{Introduction}
Learning and capturing hand-object interactions stands as a foundation for understanding human behaviors and has many practical applications. From augmented and virtual reality \cite{hurst2013gesture} to the domains of robotic grasps, human-computer interaction \cite{sridhar2015investigating, piumsomboon2013user} and learning from human demonstrations \cite{garcia2020physics, handa2020dexpilot, qin2022dexmv, wang2023event}, the comprehension and replication of these interactions are of significance. 

In this work, we focus on understanding the interactions involved in generating 3D human grasps for objects. We recognize most existing grasp generation methods depend on full 3D object models \cite{jiang2021hand, karunratanakul2020grasping}, yet in real-world scenarios, objects and scenes are commonly perceived from a single viewpoint, leading to incomplete object point clouds with potential scene noise. Consequently, existing methods find it challenging to capture the complete geometric features of a partial object and generate human grasps based on such single-view point clouds, limiting their practical applicability. Clearly, a major challenge of this task is the incompleteness of the object. If the model cannot perceive the object's overall shape, the generated grasp is likely to penetrate into the invisible parts of the object. Furthermore, due to interference from scene points, there is a risk that the generated hand will collide with the scene elements.


To address the challenges, we introduce \textbf{S2HGrasp}, a novel framework designed for generating human grasps on single-view scene point clouds. S2HGrasp incorporates a Global Perception module that facilitates the global understanding of partial object point clouds, and a DiffuGrasp module aiming at generating high-quality human grasps from complex inputs that contain numerous scene points. 

Specifically, the Global Perception module employs a multi-task learning paradigm, including single-view point cloud completion and classification tasks. This module helps to capture the object's global geometric features, and effectively prevents the generated hands from penetrating into the object's invisible parts. During testing, we can directly extract global features from the input single-view point clouds without the need of point cloud completion and classification tasks. Through feature alignment in latent space, our module can capture global features from the partial object in a single-stage process. Concurrently, the DiffuGrasp module is designed to generate human grasps and it adds noise to the normalized hand parameters when training. Then it conditions on the scene features and predicts the original hand parameters. In testing, the DiffuGrasp module progressively denoises a random noise with the condition of scene features, and predicts the final reasonable hand parameters. Leveraging its powerful conditional generation capabilities, the DiffuGrasp module is capable of generating high-quality grasps that are close to the object, even when the input includes scene points.

In addition, we introduce a novel dataset named \textbf{S2HGD}, specifically for human grasp generation on single-view scene point clouds. Constructed using the BlenderProc \cite{denninger2019blenderproc} simulator and derived from OakInk dataset \cite{yang2022oakink}, S2HGD has a collection of 1,668 unique objects. Each object is individually placed on a tabletop setting. RGB-D images are captured across 36 distinct viewpoints, and corresponding single-view point clouds are derived from depth information. Overall, S2HGD comprises roughly 99,000 point clouds, each paired with a reasonable human grasp.


Our experimental results indicate that our end-to-end approach S2HGrasp outperforms existing generation methods and two-stage methods without data preprocessing and TTA (test-time adaptation, which fails to work well in our task due to the incompleteness of the objects). Additionally, our model not only generates natural and plausible grasps but also exhibits the ability to generalize to unseen objects.



In summary, our main contributions are as follows: 1) We explore a new task of human grasp generation on single-view scene point clouds and achieve favorable results. 2) We propose a novel model named S2HGrasp, which employs a Global Perception module to endow the model with the ability to globally perceive single-view point clouds, and a DiffuGrasp module to generate high-quality grasps despite the complex input that contains the object and scene points. 3) We construct S2HGD, a new dataset for human grasp generation based on single-object single-view scene point clouds. Our code and dataset will both be available.

\section{Related Work}

\textbf{Hand-object interaction and grasp.} Hand-object interaction is currently receiving increasing attention and involves two main tasks: 1) reconstructing 3D models of hands and objects based on images \cite{oberweger2015hands, madadi2017end, hasson2019learning, hasson2021towards, tekin2019h+, kokic2019learning, hasson2020leveraging, doosti2020hope, yang2021cpf, cao2021reconstructing} and 2) predicting how to grasp an object based on its visual representation (images or point clouds) \cite{jiang2021hand, ye2023affordance, karunratanakul2020grasping, corona2020ganhand, taheri2020grab, brahmbhatt2019contactgrasp, li2022hgc, chen2023tracking}. GraspTTA \cite{jiang2021hand} can generate human grasps based on the consistency between hand contact points and object contact regions given complete objects. Affordance Diffusion \cite{ye2023affordance} employs diffusion model to generate hand-object interacting images given object images. When it comes to dexterous hand grasp, \cite{xu2023unidexgrasp} and \cite{wan2023unidexgrasp++} work well using generative models and reinforcement learning, and \cite{li2022hgc} use a single-shot network to predict dexterous hand grasp in clutter. Existing works mainly focus on generating grasps for complete objects, however, in real world, objects are often observed from a single viewpoint. Thus we study human grasp generation on single-view scene point clouds, employing two modules to tackle the challenges of incompleteness of the object and impact of scene elements.

\vspace{1ex}\noindent\textbf{3D point cloud completion.} 3D point cloud completion enables the model to have global perception capability and has various methods, including point-based \cite{liu2020morphing, yuan2018pcn, wen2020point}, view-based \cite{hu20203d, zhang2021view}, convolution-based \cite{xie2020grnet}, graph-based \cite{pan2020ecg}, generative model-based \cite{huang2020pf}, transformer-based \cite{yu2021pointr} approaches, etc. Point Completion Network (PCN) \cite{yuan2018pcn} is a learning-based shape completion method, and its decoder designed for point cloud features does not require any assumptions about the shape or category of the point clouds. It can generate both coarse and fine-grained completions with a relatively small number of parameters. Motivated by PCN \cite{yuan2018pcn}, we design a Global Shape Perception (GSP) to complete partial point clouds. Compared to other SOTA methods for point cloud completion, this approach allows for the presence of scene points and achieves a satisfactory completion result, giving the model global awareness.

\begin{figure*}[ht]
    \begin{center}
        \includegraphics[width=1\textwidth]{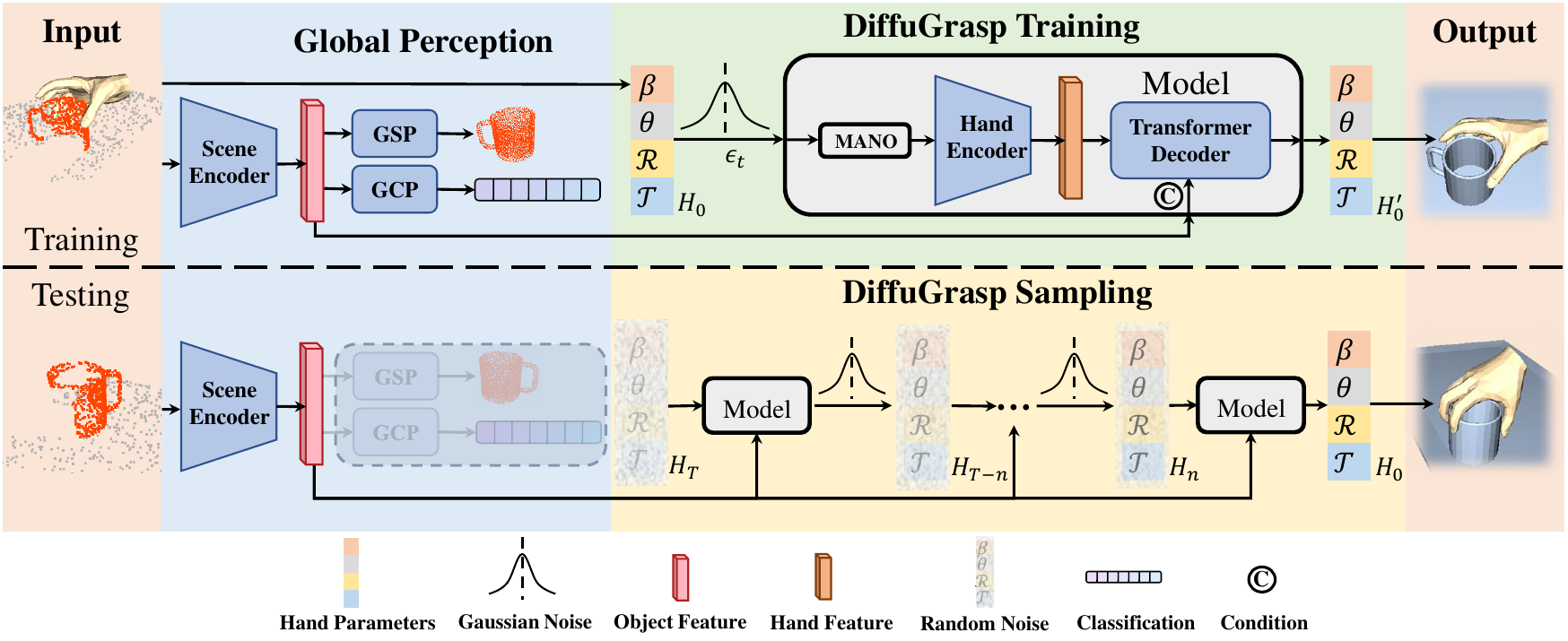}
    \end{center}
    \vspace{-15pt}
    \caption{\textbf{S2HGrasp framework}. The scene encoder takes single-view scene point clouds as input and extracts their features through PointNet++ \cite{qi2017pointnet++} and a transformer block. The features are then used in point cloud completion (GSP), classification (GCP), and grasp generation (DiffuGrasp). GSP and GCP won't be used in testing. In the DiffuGrasp Training, the model adds noise to the normalized hand parameters and extracts hand features after passing the parameters into the MANO layer \cite{romero2022embodied}. Then the object features and hand features will be fed into a transformer decoder to predict the original hand parameters. When testing, the DiffuGrasp Sampling starts from a random noise and iteratively denoises it, resulting in final hand parameters.}
    \vspace{-10pt}
    \label{framework}
\end{figure*}

\vspace{1ex}\noindent\textbf{Diffusion model.} Diffusion models have emerged as the new deep generative models that progressively destruct data by injecting noise, then learn to reverse this process for sample generation. They have been used in a variety of domains, like computer vision \cite{baranchuk2021label, cai2020learning, ho2022cascaded, zheng2023diffuvolume}, natural language processing \cite{austin2021structured, li2022diffusion}, temporal data modeling \cite{chen2020wavegrad} and multi-modal modeling \cite{avrahami2022blended}. Notably, \cite{huang2023diffusion} introduces a SceneDiffuser for 3D scene understanding, and it also generates dexterous hands by adding noise to hand parameters and denoising them. 
Differently, we convert noised hand parameters into point clouds, extract their features, and then decode them to get final grasps. This improves the naturalness of generated hands, as demonstrated later in our experiments.

\section{S2HGrasp} 
\subsection{Problem Statement and Method Overview}
We explores a novel task: generating physically plausible human grasps for objects based on single-view scene point clouds. We enable the model to globally perceive partial objects and generate high-quality grasps despite interference from scene points. The input is point clouds $\mathcal{P}\in\mathbb{R}^{N \times3}$ (where $N$ is the number of points including tabletop and partial object), captured from a single viewpoint. The output is human grasp parameters $\mathcal{H} \in \mathbb{R} ^ {61}$ modeled via differentiable MANO layer \cite{romero2022embodied}, with shape parameter $\beta \in \mathbb{R} ^ {10}$ for size of the hand and pose parameter $\theta \in \mathbb{R} ^ {45}$ for joint angle of 15 joints, as well as $\mathcal{R} \in \mathbb{R} ^ {3}$ and $\mathcal{T} \in \mathbb{R} ^ {3}$ for rotation and translation of wrist joint respectively. Given the hand parameters, MANO layer will output the hand mesh with $\hat{\mathcal{M}} = (\hat{\mathcal{V}} \in \mathcal{R}^{778 \times 3}, \hat{\mathcal{F}})$, where $\hat{\mathcal{V}}$ and $\hat{\mathcal{F}}$ denote hand vertices and faces. This task is more aligned with real-world scenarios and is beneficial for practical applications.

The generated grasps should be not only natural and plausible, but also able to firmly hold objects in physics-based simulators. Due to the incompleteness of object point clouds, the model must possess the capability to globally perceive the object to prevent hands from penetrating into the invisible parts of the object. Moreover, the model needs to generate high-quality human grasps despite interference from scene points. Therefore, we propose a model named \textbf{S2HGrasp}, as summarized in \cref {framework}, which consists of two modules: Global Perception module and DiffuGrasp module, to generate human grasps on single-view point clouds.

S2HGrasp is an end-to-end framework. The Global Perception module equips the model with the ability to perceive global geometric features of partial objects, while the DiffuGrasp module is responsible for resisting interference from scene points and generating high-quality human grasps. In the training process, our model takes single-view point clouds and corresponding hand parameters as input and uses an encoder to extract scene features. The Global Perception module endows these features with the global perception of the object. The DiffuGrasp module conditions on these features, adds noise to hand parameters, and trains a decoder to recover the original hand parameters. In testing, DiffuGrasp starts by sampling a random noise and progressively denoises it to derive the final hand parameters.

\subsection{Global Perception}
The Global Perception module aims at enabling the model to perceive the global geometric characteristics of the object from single-view point clouds. Thanks to the multi-task learning paradigm in training, our model can directly extract the scene features that contain global perception of the partial object in a single-stage process when testing. The resulting scene features will be used in the DiffuGrasp module, making the generated human grasps more natural, avoiding penetration with invisible parts of the object. 
Our scene point cloud features $\mathcal{F} _ s$ are primarily extracted using PointNet++ \cite{qi2017pointnet++}, and we utilize a transformer block similar to \cite{zhao2021point} to add attention mechanisms, making the model focus more on the local regions suitable for grasping. 

Firstly, we design a Global Shape Perception (GSP) to perform point cloud completion task on the extracted feature, motivated by PCN \cite{yuan2018pcn}. After getting the feature $\mathcal{F}_s$ of single-view scene point clouds, we pass it through our GSP and result in $\mathcal{N}$ points of the completed object. Subsequently, containing the global perception of the object, $\mathcal{F}_s$ can be better utilized in the human grasp generation. 


Secondly, we design a Global Category Perception (GCP) to perform point cloud classification task. Specifically, we feed the obtained scene feature $\mathcal{F}_s$ into a classification head to output the object category. Since the object is incomplete and the input contains both the object and scene points, this classification task is challenging. Nevertheless, we calculate the F1 score for the classification results of S2HGrasp trained on S2HGD, achieving a score of 0.96. It indicates that GCP performs well, enabling the model to focus more on the object. By predicting the object's category, it can better perceive the object's overall shape. 

\noindent \textbf{Global Perception Loss}. In the GSP of the Global Perception module, we use a Chamfer Distance loss $L _ {GSP}$ between the completed point clouds and the corresponding complete object point clouds for supervision, enabling it to have the capability to perceive the global feature of the object. In the GCP, we use Cross Entropy Loss $L _ {GCP}$. The total loss of the Global Perception module is defined as,
\begin{equation} \label{eq:global_perception_loss}
\setlength{\abovedisplayskip}{3pt}
\setlength{\belowdisplayskip}{3pt}
    L_{Glo} = \lambda_{GSP} \cdot L_{GSP} + \lambda_{GCP} \cdot L_{GCP}.
\end{equation}

\subsection{DiffuGrasp}
Since the input for our task is single-view scene point clouds including the tabletop and the object, the presence of numerous scene points can affect the quality of the generated hand and may lead to collision with the scene points. Therefore, we design the DiffuGrasp module with two innovations, to generate high-quality and diverse human grasps from complex input. Besides, we also design a plane loss to avoid collision between the hand and the tabletop.

First, we add noise to hand parameters, but direct noise injection is not feasible due to the differences in the ranges and meanings of the parameters. Therefore, we perform specific normalization for each parameter based on its meaning and range. Second, hand parameters include hand shape, pose, and position information, and directly decoding parameters does not effectively model these information. Thus, we pass them through MANO layer \cite{romero2022embodied} to convert them into point clouds and use hand encoder to obtain their features before feeding them into the decoder.

Following the practices of conditional diffusion model \cite{ho2022cascaded, rombach2022high}, in our work, the DiffuGrasp module is with condition $\mathcal{F} _ s$, \textit{i.e.}, the feature of single-view point clouds. The DiffuGrasp module can progressively generate human grasps that are close to the object and avoid penetration with the object and the scene during the denoising process. 

\noindent \textbf{- Training Process}. Conditioning on the single-view scene point cloud feature $\mathcal{F} _ s$, the DiffuGrasp module adds noise to the normalized hand parameters $\mathcal{H} \in \mathbb{R} ^ {61}$. After getting the hand feature $\mathcal{F} _ h$, we feed it together with $\mathcal{F} _ s$ into a transformer decoder and train the decoder at various noise levels to predict the original hand parameters, \textit{i.e.}, denoise $\mathcal{H}_0$ for different $\mathcal{H}_t$ given $t$, shown as below:
\begin{equation}
\setlength{\abovedisplayskip}{3pt}
\setlength{\belowdisplayskip}{3pt}
    \mathcal{H}_t = \sqrt{1-\overline{\alpha}_t} \mathcal{H}_0 + \sqrt{\overline{\alpha}_t} \cdot \epsilon, 
    \mathcal{H}'_0 = \operatorname{Model} (\mathcal{H}_t),
\end{equation}
where $ \epsilon$ is noise, $\alpha$ is noise coefficient and $ \overline{\alpha}_t = \prod_{i=1}^t \alpha_i$.




\noindent \textbf{- DiffuGrasp Loss}. Referring to the codes in DDPM \cite{ho2020denoising}, calculating the loss on the input and calculating the loss on the noise are equivalent, so we use \textit{Reconstruction Loss} along with \textit{Cmap Loss}, \textit{Penetration Loss} and \textit{Plane Loss} to train our transformer decoder. Specifically, 
1) \textit{Reconstruction Loss} of hand parameters and hand point clouds $\mathcal{P}^h \in \mathcal{R}^{778 \times 3}$ is used to minimize reconstruction error,
\begin{equation}
\setlength{\abovedisplayskip}{3pt}
\setlength{\belowdisplayskip}{3pt}
    L_{param} = |\mathcal{H}'_0 - \mathcal{H}_0|,
    L_{\mathcal{V}} = ||\hat{\mathcal{V}} - \mathcal{P}^h||_2^2.
\end{equation}

\noindent 2) \textit{Cmap Loss} is used to ensure the generated hand is as close to the object as possible. We compute the distance from the hand's contact points $\mathcal{V}^p$ to the object's point clouds. For each point $\mathcal{P}_i^o$, the distance $\textbf{D}(\mathcal{P}_i^o) = \operatorname{min}_j || \mathcal{V}^p_j - \mathcal{P}^o_i||_2^2$. If the distance is less than a threshold, we consider the point as a potentially contactable point on the object, \textit{i.e.}, contact map. One of our training objectives is to make the contact points on the hand as close as possible to the contact map on the object as, 
\begin{equation}
\setlength{\abovedisplayskip}{3pt}
\setlength{\belowdisplayskip}{3pt}
    L_{cmap} = \sum\nolimits_i \textbf{D}(\mathcal{P}_i^o), \text{\textit{ for all }} \textbf{D}(\mathcal{P}_i^o) \leq \mathcal{T}.
\end{equation}
\noindent 3) \textit{Penetration Loss} is used to penalize cases where the hand penetrates into the object. Specifically, we define the subset of object points that penetrate into the hand's mesh as $\mathcal{P}^o_{in}$, and the penetration loss is defined as minimizing the distances from these points to their nearest points on the hand,
\begin{equation}
\setlength{\abovedisplayskip}{3pt}
\setlength{\belowdisplayskip}{3pt}
    L_{penetr} = \frac{1}{|\mathcal{P}^{o}_{in}|} \sum_{p \in \mathcal{P}^{o}_{in}} \operatorname{min}_i ||p - \hat{\mathcal{V}_i}||_2^2.
\end{equation}

Note that we don't use the complete object to calculate the Cmap loss and the Penetration loss, and it is only used in the GSP module. Instead, we combine the input single-view point clouds $\mathcal{P}^s$ with the corresponding completion result $\mathcal{P}^c$ from the GSP for loss calculations. Specifically, we first calculate the minimum distance $\textbf{D}(\mathcal{P}_i^{c}) = \operatorname{min}_j || \mathcal{P}^s_j - \mathcal{P}^c_i||_2^2$, and keep the points $ \hat{\mathcal{P}^{c}}= \{\mathcal{P}^c_i \in \mathcal{P}^c \mid \textbf{D}(\mathcal{P}_i^{c}) \geq \theta \}$. We get the final completion result $\mathcal{P}^{C}= \hat{\mathcal{P}^{c}} \cup \mathcal{P}^s$, which is used for loss calculation. This approach prevents large errors from directly using rough completion results to compute loss functions. It also allows the model to have a global perception of the object's shape and avoids generating hands that penetrate into the unseen parts of the object. 

\noindent 4) \textit{Plane Loss}. As for generating human grasps based on scene point clouds, one of the challenges is to avoid contact between the hand and the tabletop point clouds. Thus, we design a plane loss to penalize cases where there is penetration between the tabletop point clouds and the hand. To avoid the enormous computational cost of calculating distances between tabletop point clouds and hand vertices, we regress the parameters of the tabletop plane. We define the hand points which are on the backside of the plane as $\hat{\mathcal{V}}_{back}$, and calculate the distance $\textbf{D}(\hat{\mathcal{V}}_i, P)$ from each point $\hat{\mathcal{V}}_i$ in $\hat{\mathcal{V}}_{back}$ to the plane $P$, and the plane loss is defined as,  
\begin{equation}
\setlength{\abovedisplayskip}{3pt}
\setlength{\belowdisplayskip}{3pt}
    L_{plane} = \sum\nolimits_i \textbf{D}(\hat{\mathcal{V}}_i, P), \text{\textit{ for all }} \hat{\mathcal{V}}_i \in \hat{\mathcal{V}}_{back}.
\end{equation}
Thus, the loss for the DiffuGrasp is:
\begin{equation}
\setlength{\abovedisplayskip}{3pt}
\setlength{\belowdisplayskip}{3pt}
\begin{aligned}
    L_{Diff} = \lambda_{param} \cdot L_{param} + \lambda_\mathcal{V} \cdot L_\mathcal{V} + \\
    \lambda_{c} \cdot L_{cmap} + \lambda_{pe} \cdot L_{penetr} + \lambda_{pl} \cdot L_{plane}
\end{aligned}
\end{equation}



\noindent \textbf{- Sampling Process.} The Sampling Process begins by randomly sampling a noise from standard Gaussian distribution. Subsequently, it undergoes a denoising process for $T$ timesteps with the feature of single-view point clouds serving as a condition, and ultimately yields the final hand parameters. The normalization follows the same procedure as described in the Training Process. We add noise calculated from $\mathcal{H}_t$ and $t$ to hand parameters $\mathcal{H}_t$, and then feed them into the \textbf{Model}, which includes the MANO layer, Hand Encoder, and the Transformer Decoder as shown in \cref {framework}, to get $\mathcal{H}_{t-1}$. After $T$ iterations of the denoising process, the model will get final $\mathcal{H}'_0$. The distribution $p_\theta\left(\mathcal{H}_0 | \mathcal{F} _ s\right)$ represents our sampling process and can be shown as,
\begin{align}
    p_\theta\left(\mathcal{H}_0 | \mathcal{F} _ s\right) & = p\left(\mathcal{H}_T \right) \prod_{t=1}^T p\left(\mathcal{H}_{t-1}|\mathcal{H}_t, \mathcal{F} _ s\right), \\
    p(\mathcal{H}_{t-1} | \mathcal{H}_{t}, \mathcal{F} _ s) & = 
    \mathcal{N}(\mathcal{H}_{t-1}; \mu_\theta (\mathcal{H}_t, t, \mathcal{F}_s), \sum_\theta(\mathcal{H}_t, t, \mathcal{F} _ s ) )
\end{align}
\vspace{-10pt}
\begin{equation}
\setlength{\abovedisplayskip}{3pt}
\setlength{\belowdisplayskip}{3pt}
    \mathcal{H}_{t-1} = \sqrt{\overline{\alpha}_{t-1}} \mathcal{H}_0 + \sqrt{1 - \overline{\alpha}_{t-1} - \sigma_t^2} \cdot \epsilon_\theta^{(t)}(\mathcal{H}_{t}) + \sigma_t\epsilon_t,
\end{equation}
where $\sigma_t = \eta \sqrt{\frac{1- \overline{\alpha}_{t-1}}{1- \overline{\alpha}_t} \cdot (1 - \frac{ \overline{\alpha}_t}{\overline{\alpha}_{t-1}})}$, $\eta$ is DDIM \cite{song2020denoising} sampling coefficient and $\epsilon_t \sim \mathcal{N}(\textbf{0}, \textbf{\textit I})$ is standard Gaussian noise. For more details, please refer to supplementary file.

\section{S2HGD Dataset}
To better align with real-world scenarios where objects are typically observed from a single viewpoint and to further explore generating human grasps based on single-view scene point clouds, we construct the \textbf{S2HGD} dataset. Here we show how we build our S2HGD with BlenderProc \cite{denninger2019blenderproc}: 

\vspace{1ex}\noindent\textbf{Data Source.} Our raw data (including objects and human grasp annotations) all come from the OakInk dataset \cite{yang2022oakink}. We select a total of 1,667 objects from 35 different categories (\textit{e.g.} bottle, knife, mug) as the dataset's objects.

\vspace{1ex}\noindent\textbf{Simulation Environment Setting.} In the simulation environment, we select a table from ShapeNet \cite{chang2015shapenet} and position 36 cameras above the table, arranged in three concentric circular arrays at heights of 0.25 meters, 0.45 meters, and 0.65 meters from the table respectively. Each circular array contains 12 cameras at the same height, with a 30-degree spacing between each camera, as shown in \cref{dataset and results}.

\vspace{1ex}\noindent\textbf{Data Processing.} We drop an object from 20 $cm$ above the table, allowing it to fall onto the tabletop freely. This is repeated ten times to get ten different poses. For each pose, we capture an RGBD image from each camera and convert it into single-view point clouds. After collecting all point clouds, we test all grasps of each object, discard any that contact the table and remove scenes lacking collision-free grasp annotations. From the remaining grasps, we choose the one closest to the object as the final annotation.

To enable our model to generate high-quality grasps for both varying viewpoints of seen objects and entirely unseen objects, we partition our dataset in two unique ways: \textbf{View-S2HGD} focuses on varying viewpoints, whereas \textbf{Object-S2HGD} is based on object diversity. Specifically, View-S2HGD includes different single-view point clouds of the same objects in training and testing sets. Object-S2HGD ensures that the objects in the testing set are entirely absent from the training set, allowing us to assess the model's capacity for generalization to unseen objects.

In total, S2HGD contains 1,667 objects and about 99,000 single-view scene point clouds with one annotation each.

\begin{figure*}[ht]
    \begin{center}
        \includegraphics[width=1\textwidth]{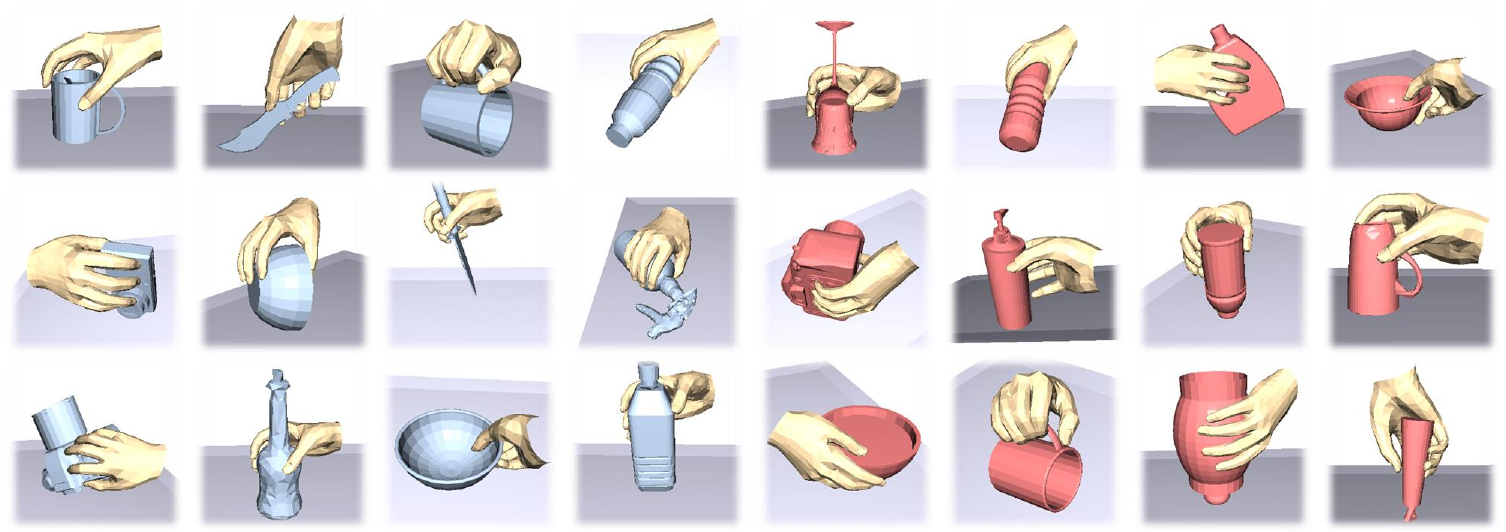}
    \end{center}
    \vspace{-15pt}
    \caption{The visualization of generated grasps of our S2HGrasp on our two datasets. The \textcolor{color4}{blue} objects on the left represent the results of View-S2HGD, while the \textcolor{color3}{red} objects on the right represent the results of Object-S2HGD.}
    \vspace{-10pt}
    \label{result}
\end{figure*}

\begin{table*}[ht] \small
\centering

\scalebox{0.9}{
\begin{tabular}{ll|ccc|ccc||cc|cc}

\multicolumn{2}{c|}{\multirow{2}{*}{}}                          & \multicolumn{3}{c|}{View-S2HGD} & \multicolumn{3}{c||}{Object-S2HGD} & \multicolumn{2}{c|}{HO3D-object} & \multicolumn{2}{c}{Obman-object} \\
\multicolumn{2}{c|}{}                                           & GT        & GC \cite{jiang2021hand} & Ours     & GT        & GC \cite{jiang2021hand}       & Ours      & GC \cite{jiang2021hand}         & Ours        & GC  \cite{jiang2021hand}        & Ours        \\ \hline
\multirow{2}{*}{Penetration}    & Depth($cm$) $\downarrow$     & 0.27 & 0.23 & \colorbox{color2}{\textbf{0.21}} & 0.26 & 0.25 & \colorbox{color2}{\textbf{0.21}} & 2.11 & \textbf{1.38} & 0.32 & \textbf{0.21}        \\
                                & \colorbox{color1}{\textbf{Volume(${cm}^3$)$\downarrow$}} & 6.60 & 13.54  & \colorbox{color2}{\textbf{6.58}} & 7.22  & 14.48 & \colorbox{color2}{\textbf{5.58}}  & 8.66  & \textbf{4.87}  & 9.84  & \textbf{4.62}       \\ \hline
\multirow{2}{*}{Grasp Displace} & \colorbox{color1}{\textbf{Mean($cm$)$\downarrow$}}      & 2.01 & 3.10 & \textbf{2.73} & 2.18 & 3.69 & \textbf{3.26} & 3.61 & \textbf{3.30} & 4.18 & \textbf{3.25}        \\
                                & Variance($cm$) $\downarrow$  & $\pm$2.73 & $\pm$3.49 & \textbf{$\pm$3.16} & $\pm$2.69 & $\pm$3.94 & \textbf{$\pm$3.38} & $\pm$3.93 & \textbf{$\pm$3.73} & $\pm$4.31 & \textbf{$\pm$3.75}        \\ \hline
Perceptual Score                & \{1,…,5\} $\uparrow$         & 4.54    & 2.60    & \textbf{3.25}  & 4.52    & 2.54    & \textbf{3.14}  & 2.11    & \textbf{2.89}    & 2.04    & \textbf{3.13}           \\ \hline
Contact                         & Ratio(\%)$\uparrow$          & 100  & 98.07  & \textbf{99.41}  & 100  & 97.87  & \textbf{98.67}  & 97.62  & \textbf{99.06}  & 85.82  & \textbf{96.06}         \\ \hline
\multirow{2}{*}{Diversity($\sigma^2$)} & Axis($10^{-2}$) $\uparrow$           & 2.15 & 0.85 & \textbf{1.03} & 2.29 & 0.52 & \textbf{0.71} & 0.50 & \textbf{0.79} & 0.56 & \textbf{0.70}        \\
                                & Angle($10^{-2}$) $\uparrow$          & 8.86 & 3.29 & \textbf{4.51} & 9.68 & 2.04 & \textbf{2.75} & 1.32 & \textbf{1.95} & 2.28 & \textbf{2.67}        \\
\end{tabular}
}

\caption{Quantitative results compared with GraspCVAE \cite{jiang2021hand} on S2HGD. The objects in HO3D and Obman datasets are used to demonstrate the generalization capability. The \textbf{Penetration Volume} metric and the \textbf{Grasp displace} metric are our main indicators. The results show that our method outperforms GraspCVAE on all indicators (in bold) and is significantly effective in reducing hand-object penetration caused by incomplete point clouds. The Penetration results of our method are even \textbf{better than GT} (with a blue background).}
\vspace{-10pt}
\label{main experiments results}
\end{table*}

\section{Experiments}

\subsection{Datasets and Experiment Setup}
\textbf{Implementation Details.} The single-view point clouds in our dataset consist of 2,000 points, with 1,000 points belonging to the object and 1,000 points to the tabletop. In training, we use Adam optimizer and LR = $1e-4$, where the LR is reduced by half when the model trained 100, 160 and 180 epochs. We totally train for 200 epochs and the batch size is 60. Before training the entire model with View-S2HGD and Object-S2HGD, we pre-train the GSP module using single-view point clouds and their corresponding complete objects from each dataset respectively for 200 epochs. The loss weights we use are $\lambda_{GSP} = 5$, $\lambda_{GCP} = 2$, $\lambda_{param} = 300$, $\lambda_{V} = 15$, $\lambda_{cmap} = 300$, $\lambda_{pe} = 15$, $\lambda_{pl} = 1$ for View-S2HGD. For Object-S2HGD, only $\lambda_{pe}$ and $\lambda_{param}$ differ, being set to $20$ and $250$, respectively. 


\noindent \textbf{Datasets.} We use S2HGD, Obman \cite{hasson2019learning} and HO3D \cite{hampali2020honnotate} for our experiments. S2HGD is the dataset we mainly conduct our experiments on. We train and test our model on both View-S2HGD and Object-S2HGD to demonstrate the overall performance. We also conduct ablation experiments on View-S2HGD. Obman and HO3D are used for studying hand-object interaction and Obman is synthetic, while HO3D is real. To validate generalization capability of our method on unseen objects, we utilize the Object-S2HGD for training and testing. We also select 30 objects from Obman and 10 objects from HO3D to construct two testing sets using the same methodology employed in the construction of S2HGD, to further assess generalization capability.

\begin{figure*}[htbp]
    \centering
    \begin{minipage}[t]{0.65\textwidth}
    \centering
        \includegraphics[width=11cm]{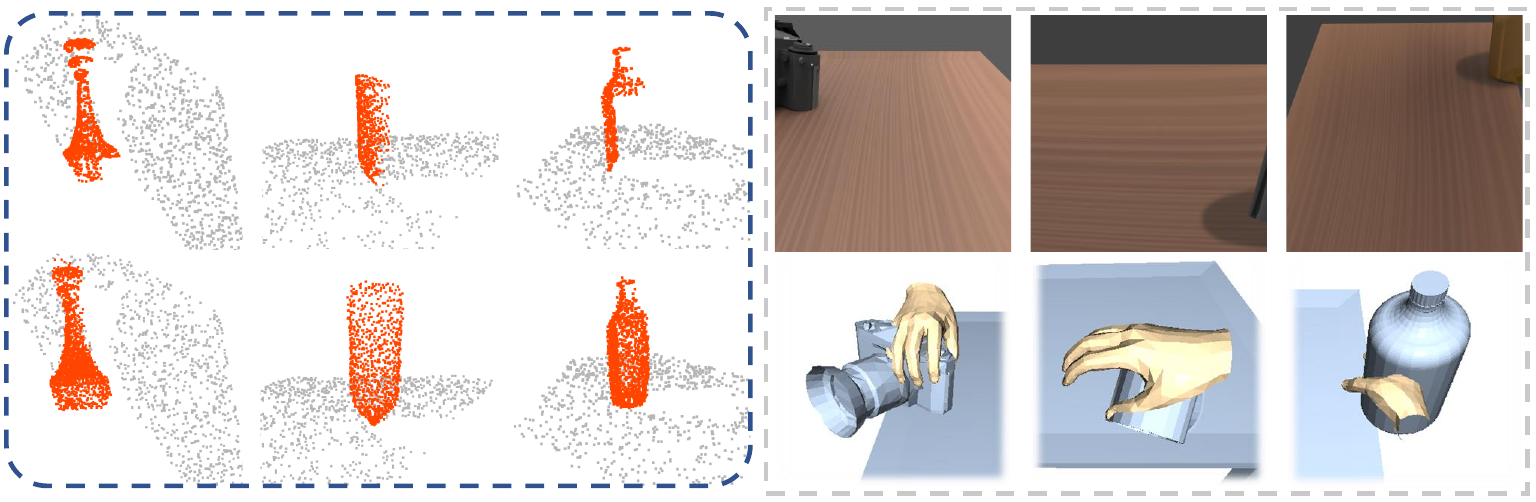}
        \caption{Visualization of point cloud completion and our failure cases. \textbf{Left}: Point cloud completion results, with the input single-view point clouds in the top row and completion results below (gray points for the tabletop, red for the object). \textbf{Right}: Failure cases, with single-view images at the top and corresponding failure cases of our method below.} 
        \label{pcn fail}
    \end{minipage}
    \hspace{0.2cm}
    \begin{minipage}[t]{0.32\textwidth}
    \centering
        \includegraphics[width=5.4cm]{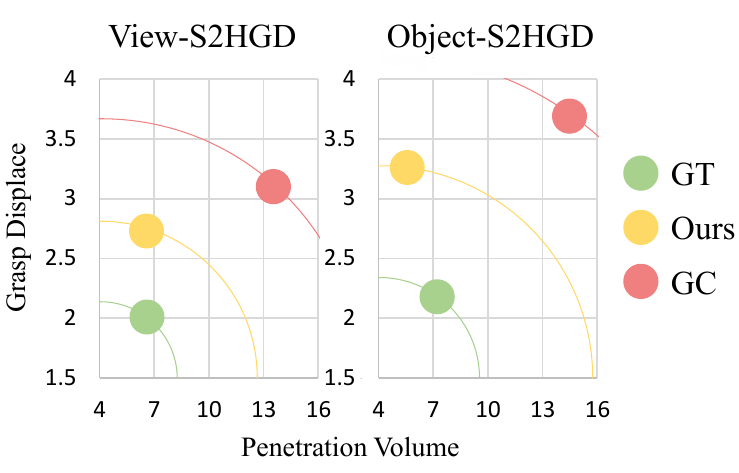}
        \caption{The balance between penetration volume (x-axis) and grasp stability (y-axis). Method whose result point is closer to the origin is more effective.}
        \label{tradeoff}
    \end{minipage}
    \vspace{-10pt}
\end{figure*}


\noindent \textbf{Evaluation Metrics.} The primary goal of our model is to generate plausible and natural human grasps for objects based on single-view scene point clouds. Therefore, our main evaluation metrics involve calculating the \textit{penetration} depth and volume between the object and the hand. We also use a physics-based simulation environment to calculate the \textit{grasp displace} to assess the stability of the hand grasping the object. \textit{Perceptual score} is used to evaluate the naturalness of the generated grasps and we invite 50 participants to rate the naturalness of generated grasps, using a scale from 1 to 5, where higher scores indicate higher naturalness. What's more, to demonstrate the contact between the generated hand and the object, we also calculate \textit{contact ratio} between the hand and the object. Besides, to demonstrate diversity of generated grasp poses, we introduce a \textit{diversity} metric. Specifically, we calculate the variance of the rotation axes $\sigma_{axis}^2$ and angles $\sigma_{angle}^2$ for 15 joints, excluding the wrist joint, across all grasp samples. Details about the evaluation metrics can be found in the supplementary file.


\subsection{Grasp Generation Performance}


\textbf{Visualization results.} We visualize the generated grasps on both the View-S2HGD and Object-S2HGD datasets, as shown in \cref {result}. The results demonstrate that our model can generate natural and plausible human grasps for different single-view point clouds of seen objects and shows a notable level of generalization for unseen objects. We also visualize the performance of our GSP module, \textit{i.e.} the partial object point cloud completion results, as shown in the left of \cref {pcn fail}. The first row is the input single-view point clouds and the second row is the completion results. The output of our GSP module consists of only the red points. We also visualize some failure cases, shown in the right part of \cref{pcn fail}. The first example fails due to the complex shape of the object and it's difficult to perceive its global shape from only partial point clouds, leading to the generated hand penetrating into the object. The last two examples fail because only a small part of the object is visible from the viewpoint. There are too few points to generate a reasonable grasp. More visualization results are in the supplementary file, including the HO3D and Obman datasets.

\begin{table}[t] \small
\centering

\setlength{\tabcolsep}{2.5pt}
\begin{tabular}{ccccc}
\toprule[1pt]
 \multirow{2}{*}{Method} & \multicolumn{2}{c}{Penetration} & Grasp Displace & Contact  \\
 &  Depth$\downarrow$ & Volume$\downarrow$ & Mean±Variance$\downarrow$  & Ratio$\uparrow$ \\ \hline
 w/o encoder \cite{huang2023diffusion} & 0.21      & 6.70                          & 3.13$\pm$3.40       & 98.49       \\
 w/ encoder(ours)     & 0.21      & \textbf{6.58}                 & \textbf{2.73$\pm$3.16}      & \textbf{99.41}       \\ 
\bottomrule[1pt]
\end{tabular}
\vspace{-5pt}
\caption{Comparison with method \cite{huang2023diffusion} that without hand encoder.}
\vspace{-15pt}
\label{scene}
\end{table}

\noindent \textbf{Comparing with GraspCVAE.} As shown in Table \ref{main experiments results}, we conduct experiments on our View-S2HGD and Object-S2HGD datasets and the testing sets we construct using the objects selected from HO3D \cite{hampali2020honnotate} and Obman \cite{hasson2019learning} datasets. We compare our model with the SOTA method GraspTTA \cite{jiang2021hand} for the task of generating human grasps based on complete object point clouds. GraspTTA employs ContactNet for Test-Time Adaptation (TTA), which predicts contact areas with the condition of complete object point clouds. Since our task involves single-view scene point clouds as input, with partial object point clouds and interference from scene point clouds, TTA is challenging to apply. Thus, in the comparative experiments, we did not include the TTA module, only utilizing the GraspCVAE network (GC). 

\begin{table}[t] \small
\centering

\setlength{\tabcolsep}{3pt}
\begin{tabular}{ccccc}
\toprule[1pt]
 \multirow{2}{*}{Method} & \multicolumn{2}{c}{Penetration} & Grasp Displace & Contact  \\
 &  Depth$\downarrow$ & Volume$\downarrow$ & Mean±Variance$\downarrow$  & Ratio$\uparrow$ \\ \hline
 PCN+GC & 0.22      & 11.34                          & 6.00$\pm$4.83      & 85.66       \\
 PCN+DG & 0.20      & 7.68                          & 6.50$\pm$4.51      & 72.85       \\
 Ours          & 0.21      & \textbf{5.58}                           & \textbf{3.26$\pm$3.38}      & \textbf{98.67}       \\ 
\bottomrule[1pt]
\end{tabular}
\vspace{-5pt}
\caption{Comparison with two-stage methods. We use our model trained on Object-S2HGD to compare with others. The two-stage methods initially complete the single-view point cloud (by PCN \cite{yuan2018pcn}), and generate human hand grasps on the completed point cloud using GraspCVAE (GC) \cite{jiang2021hand} and our DiffuGrasp(DG).} 
\vspace{-0.5cm}
\label{two_stage_result}
\end{table}

The results indicate that our model outperforms GraspCVAE across all datasets. From the ``Penetration" metric, it can be seen that GraspCVAE, due to its lack of global perception of single-view point clouds, produces grasps that are significantly penetrate into the object, while ours are much less, even better than GT. Additionally, our model also demonstrates superior generalization capability on unseen objects compared to GraspCVAE. Regarding the ``Grasp Displace" metric, a larger penetration volume may result in better grasping stability (the larger the penetration, the greater the force on the fingers in the simulator), which could make the results of GraspCVAE seem favorable. However, an ideal grasp should have both smaller penetration volume and good stability. Therefore, we present the balance on these two metrics of both methods in \cref{tradeoff}. It is evident that the points representing our method are closer to the origin point, which also indicates that our method better unifies penetration with grasp stability, meaning the generated grasps are more natural and reasonable.
\begin{table*}[ht] \small
\centering

\begin{tabular}{cccccccc}
\toprule[1pt]
\multirow{2}{*}{Method} & \multirow{2}{*}{CVAE} & \multirow{2}{*}{DiffuGrasp} & \multirow{2}{*}{GP} & \multicolumn{2}{c}{penetration}     & Grasp Displace   & Contact  \\
                     &                        &      &                                            & Depth($cm$)$\downarrow$ & Volume(${cm}^3$)$\downarrow$ & Mean $\pm$ Variance($cm$)$\downarrow$  & Ratio(\%)$\uparrow$ \\ \hline
CVAE                 & \checkmark             &      &                                               & 0.23        & 13.54              & 3.10 $\pm$ 3.49      & 98.07 \\
DiffuGrasp &               & \checkmark               &                                              & 0.20        & 11.54              & 3.06 $\pm$ 3.33      & 98.80 \\
DiffuGrasp + GP &   & \checkmark               & \checkmark                       & \textbf{0.21}        & \textbf{6.58}              & \textbf{2.73 $\pm$ 3.16}      & \textbf{99.41} \\     

\bottomrule[1pt]
\end{tabular}
\vspace{-5pt}
\caption{Ablation results of the DiffuGrasp module and the Global Perception module(GP) on our View-S2HGD dataset. The role of DiffuGrasp is demonstrated through a comparison with CVAE. }
\vspace{-15pt}
\label{tab:ablation}
\end{table*}

To generate more natural and stable grasps, instead of directly passing hand parameters into the decoder as done in \cite{huang2023diffusion}, we first convert the hand parameters into point clouds and extract their features. We then feed the features into the decoder to predict the hand parameters. Our method incorporates more shape and positional information about the hand than direct decoding the hand parameters. To verify the effectiveness of our method, we conduct comparative experiments with the approach without the hand encoder used in \cite{huang2023diffusion}, as shown in Table \ref{scene}. The results show that, with a similar penetration volume, our method achieves higher grasp stability (smaller grasp displace), thereby effectively facilitating the generation of higher quality grasps and demonstrating the advantages of our approach.

\noindent \textbf{Comparing with two-stage methods.} For the task of generating human grasps based on single-view scene point clouds, one intuitive approach is to first utilize point cloud completion methods to complete the single-view point clouds. Subsequently, a generation method like GraspTTA \cite{jiang2021hand} is employed to generate grasps on the completed point clouds. \cite{wei2022dvgg} and \cite{lundell2021ddgc} utilize such two-stage methods to generate dexterous hand grasps in single-view scenes.

We first train a Point Completion Network (PCN) \cite{yuan2018pcn} to complete the input single-view point clouds into full objects with tabletop points. Then, we train GraspCVAE (GC) \cite{jiang2021hand} and our DiffuGrasp (DG) to generate human grasps based on these completed results. The testing results are shown in Table \ref{two_stage_result}. The results demonstrate that our method outperforms both two-stage methods by a large margin. The reason is that the two-stage method is not end-to-end, which leads to accumulated errors in the process, impacting the quality of generated grasps. In addition, most works of point cloud completion are not suitable for such a situation with scene points, which will lead to a decrease in the accuracy of point cloud completion. However, the two-stage method has very high requirements for the completion results of single-view point clouds. If the result is poor, the generated grasp can easily penetrate into the object or not make contact with it. In contrast, our end-to-end model effectively avoids such problems. Last, the overall models of two-stage methods tend to be larger, require longer and more tedious training processes, and are less straightforward to train compared to our end-to-end method.

\subsection{Ablation Study}
We conduct an ablation study on our S2HGrasp using our View-S2HGD dataset to highlight the role of the Global Perception module and demonstrate the advantages of our DiffuGrasp module over CVAE, as shown in Table \ref{tab:ablation}.


\noindent \textbf{Global Perception.} The comparative results, as well as the visualizations of point cloud completion, highlight that this module helps the object encoder capture global object characteristics. This, in turn, assists the generation models in producing more natural, collision-free human grasps. 

Specifically, by comparing the results of ``DiffuGrasp" and ``DiffuGrasp+GP", we can find that with the Global Perception module, our model can effectively reduce the volume of hand-object penetration (from $11.54$ to $6.58$) and enhance the stability of the grasp. This also demonstrates that, with the aid of the Global Perception module, our model has a good perception of the overall shape of local objects, and can effectively prevent the generated hand from penetrating into the invisible areas of the object. 

\noindent \textbf{DiffuGrasp.} The previous work GraspTTA \cite{jiang2021hand} leverages CVAE to generate human grasps based on the full object model, but suffers from severe model collapse and can't generate diverse grasps poses. We believe that the generative capabilities of CVAE are not sufficient, especially for the more challenging task of generating human grasps based on single-view scene point clouds. Therefore, we conduct comparative experiments on our View-S2HGD dataset using our DiffuGrasp and CVAE (without TTA, as TTA cannot be applied to incomplete point clouds) to demonstrate the advantages of our DiffuGrasp in this task. 

Results show that even without the Global Perception module, the penetration volume of DiffuGrasp is smaller than that of CVAE ($11.54$ to $13.54$). Moreover, DiffuGrasp achieves higher grasp stability than CVAE with smaller penetration, which also shows that the grasps generated by DiffuGrasp are more reasonable and natural. In short, despite the incompleteness of the object and the interference from numerous scene points, DiffuGrasp outperforms CVAE and can generate stable human grasps with minimal penetration into the object. Additional ablation studies and experiments are shown in the supplementary file.
\section{Conclusion}
In this work, we explore a new task of generating human grasps based on single-view scene point clouds rather than full object models. We propose S2HGrasp, along with a new synthetic dataset S2HGD, to address the problem of hand-object penetration caused by the incompleteness of object point clouds. We design a Global Perception module to globally perceive partial objects and a DiffuGrasp module to generate plausible and natural human grasps despite numerous scene points. Our S2HGrasp effectively overcomes the limitation of existing approaches and extends human grasp generation beyond full object models, benefiting the study of hand-object interactions. The experimental results demonstrate that S2HGrasp outperforms other methods and achieves satisfying performance and generalization capabilities across different datasets and unseen objects.

\raggedright
\noindent \textbf{Acknowledgments.} This work was supported partially by the National Key Research and Development Program of China(2023YFA1008503),NSFC(U21A20471,U1911401), \\Guangdong NSF Project (No. 2023B1515040025, 2020B1\\515120085).
{
    \small
    \bibliographystyle{ieeenat_fullname}
    \bibliography{main}
}


\end{document}